\documentclass[runningheads]{llncs}
\usepackage{graphicx}
\usepackage{amssymb}
\usepackage{xcolor}
\usepackage{hyperref}
\def\debug{1}

\ifnum \debug=1
\providecommand{\anote}[1]{\textcolor{red}{\textbf{[AP: #1]}}}
\providecommand{\mnote}[1]{\textcolor{violet}{\textbf{[MB: #1]}}}
\else
\providecommand{\anote}[1]{}
\providecommand{\mnote}[1]{}
\fi

\begin{document}
\title{Automatic Generation of Semantic Parts for Face Image Synthesis}
%
%\titlerunning{Abbreviated paper title}
% If the paper title is too long for the running head, you can set
% an abbreviated paper title here
%
\author{Tomaso Fontanini\orcidID{0000-0001-6595-4874} \and
Claudio Ferrari\orcidID{0000-0001-9465-6753} \and
Massimo Bertozzi\orcidID{0000-0003-1463-5384} \and
Andrea Prati\orcidID{0000-0002-1211-529X}}
\authorrunning{T. Fontanini, et al.}
% First names are abbreviated in the running head.
% If there are more than two authors, 'et al.' is used.
%
\institute{IMP Lab, Department of Engineering and Architecture, University of Parma
\email{\{tomaso.fontanini, claudio.ferrari2, massimo.bertozzi, andrea.prati\}@unipr.it}}
\maketitle              % typeset the header of the contribution
\begin{abstract}
Semantic image synthesis (SIS) refers to the problem of generating realistic imagery given a semantic segmentation mask that defines the spatial layout of object classes. Most of the approaches in the literature, other than the quality of the generated images,  put effort in finding solutions to increase the generation diversity in terms of style \textit{i.e.} texture. However, they all neglect a different feature, which is the possibility of manipulating the layout provided by the mask. 
Currently, the only way to do so is manually by means of graphical users interfaces.
In this paper, we describe a network architecture to address the problem of automatically manipulating or generating the shape of object classes in semantic segmentation masks, with specific focus on human faces. Our proposed model allows embedding the mask class-wise into a latent space where each class embedding can be independently edited. Then, a bi-directional LSTM block and a convolutional decoder output a new, locally manipulated mask. We report quantitative and qualitative results on the CelebMask-HQ dataset, which show our model can both faithfully reconstruct and modify a segmentation mask at the class level. Also, we show our model can be put before a SIS generator, opening the way to a fully automatic generation control of both shape and texture. Code available at \url{https://github.com/TFonta/Semantic-VAE}.

\end{abstract}
\section{Introduction}
%Deep learning for image generation has drawn a lot of attention recently. In particular, among the several tasks in this field, 
The task of Semantic Image Synthesis (SIS) consists in generating a photo-realistic image given a semantic segmentation mask that defines the shape of objects. The mask is usually an image in which the pixel values define a specific semantic class (like eyes, skin, hair, \textit{etc.} in the case of human face).
This allows for accurately defining the spatial layout and shape of the generated images, while maintaining a high degree of freedom in terms of textures and colors.
Indeed, those can be randomly generated~\cite{wang2021image} or by extracting a specific style from a reference image~\cite{lee2020maskgan,park2019semantic}.

A nice feature of SIS methods is that the semantic mask can be manipulated to alter the shape of objects in the generated samples. 
However, currently this is done manually by using custom painting software allowing the user to modify the shape of one or more mask parts. Attempts of performing automatic face shape parts manipulation have been done, yet with different techniques, such as by using a 3D deformable model of the face~\cite{ferrari2022makes}.
Whereas manual alteration of the semantic masks is fun, it turns out impractical when the objective is to modify the shape of a large number of images. 

In the attempt of overcoming this limitation, in this paper we explore the problem of the automatic generation and manipulation of classes in segmentation masks, and propose a method that allows to generate and edit the shape of any number of parts. The proposed model can be used to produce a large variety of novel semantic masks that can then be used in conjunction with any SIS model to generate previously unseen photo-realistic RGB images. This is achieved by designing an architecture composed by an encoder that embeds each of the semantic mask parts separately, a recurrent module composed by a series of bi-directional LSTMs~\cite{schuster1997bidirectional} that learns the relationships between the shape of different mask parts and, finally, a decoder that maps the latent representation back into a realistic semantic mask. The model is trained as a Variational Autoencoder (VAE), so combining a reconstruction loss with a KL divergence in order to induce a specific distribution in the latent space. This enables the generation, interpolation or perturbation of semantic classes; these specific features, to the best of our knowledge, are still unexplored in the literature. Overall, the main contributions of this paper are the following:
\begin{itemize}
    \item we explore the novel problem of automatic generation and editing of local semantic classes in segmentation masks, independently from the others;
    \item we propose a novel architecture combining a VAE and a recurrent module that learns spatial relationships among semantic classes by treating them as elements of a sequence, under the observation that the shape of each part has an influence on the surrounding ones. More in detail, each part embedding is subsequently fed into the LSTM block so to account for shape dependencies, and then employed by the decoder to generate the final mask. The proposed architecture can finally be used in combination with any SIS architecture to boost the shape diversity of the generated samples;
    \item we quantitatively and qualitatively validate our proposal in the task of face parts editing, and report and extensive analysis of the advantages, limitations and challenges.
\end{itemize}

\section{Related Works}
%Semantic Image Synthesis
Given that no prior works addressed the problem presented in this paper, in the following we summarize some recent literature works on semantic image synthesis and variational autoencoders.

\noindent \textbf{Semantic Image Synthesis.} Semantic Image Synthesis approaches can be divided into two main categories: diversity-driven and quality-driven. Both of them take inspiration and improve upon the seminal work of Park \textit{et al.}, named SPADE~\cite{park2019semantic}, where semantic image synthesis is achieved by means of custom, spatially-adaptive normalization layers. Methods in the former category focus on the task of generating samples having the shape conditioned over semantic masks, but the style is generated randomly in order to achieve an high degree of multi-modality. Some examples of these approaches are~\cite{richardson2021encoding,wang2021image}. The trend here points towards increasing the granularity of the generated texture; for example, in CLADE~\cite{tan2021efficient} styles are generated at the class-level, while INADE~\cite{tan2021diverse} is able to generate instance-specific style by sampling from a class-wise estimated distribution. On the other side, quality-driven methods try to extract a specific style from a target image and to apply it over the generated results, in the attempt of both maintaining the shape defined by the mask and the texture defined by a reference image. An example of paper falling in this category is MaskGAN~\cite{lee2020maskgan}, in which a style mapping between an input mask and a target image is achieved using instance normalization. Also in this case, efforts are put into finding solutions to increase the precision and granularity of the style control. To this aim, Zhu \textit{et al.} developed SEAN~\cite{Zhu_2020_CVPR}, a method that is able to extract the style class-wise from each of the different semantic part of an image and map it locally over the corresponding area of the input mask. Another work following the same trend is SC-GAN~\cite{wang2021image}. Overall, it turns out clearly that none of the recent literature works deals with the problem of locally manipulating the face shape by acting on segmentation masks.  

%VAE
\noindent \textbf{Variational Autoencoders.} Autoencoders introduced in \cite{mcclelland1987parallel} were proposed as a way to achieve a compressed latent representation of a set of data, but they lack generation capabilities. On the contrary, Variational Autoencoders (VAE) \cite{kingma2013auto} described data generation through a probabilistic distribution. Indeed, once trained using a combination of reconstruction loss and Kullback-Leibler divergence, they can generate new data by simply sampling a random latent distribution and feeding it to the decoder. There exist several variations of VAE such as Info-VAE \cite{zhao2017infovae}, $\beta$-VAE \cite{higgins2017beta} and many more \cite{davidson2018hyperspherical,van2017neural}.

\section{Network Architecture}\label{sec:architecture}
The main objective that guided the design of the model architecture is that of performing automatic manipulation and generation of semantic masks, independently for each class. A semantic segmentation mask can be represented as $C$-channel image, where each channel is a binary image containing the shape of a specific object class \textit{i.e.} $M \in [0,1]^{C \times H \times W}$. So, each pixel belongs to a unique class \textit{i.e.} has value 1 only in a single channel, and each class shape is complementary to all the others, \textit{i.e.} there is no intersection between the semantic classes. 

The challenge behind manipulating or generating a specific semantic class in a segmentation mask is that its shape, and the shape of all its surrounding classes, need to be adapted so that the above properties are maintained. At the same time, the spatial arrangement of each class have also to be realistic, since it is a scenario-dependent property. In the case of facial features, the spatial relations of the different face parts need to be preserved; as example, the nose should be mostly centered between eyes.
%For example, the spatial arrangement of internal facial features has a specific configuration, which needs to be preserved.

\begin{figure}[!t]
    \centering
    \includegraphics[width=0.99\linewidth]{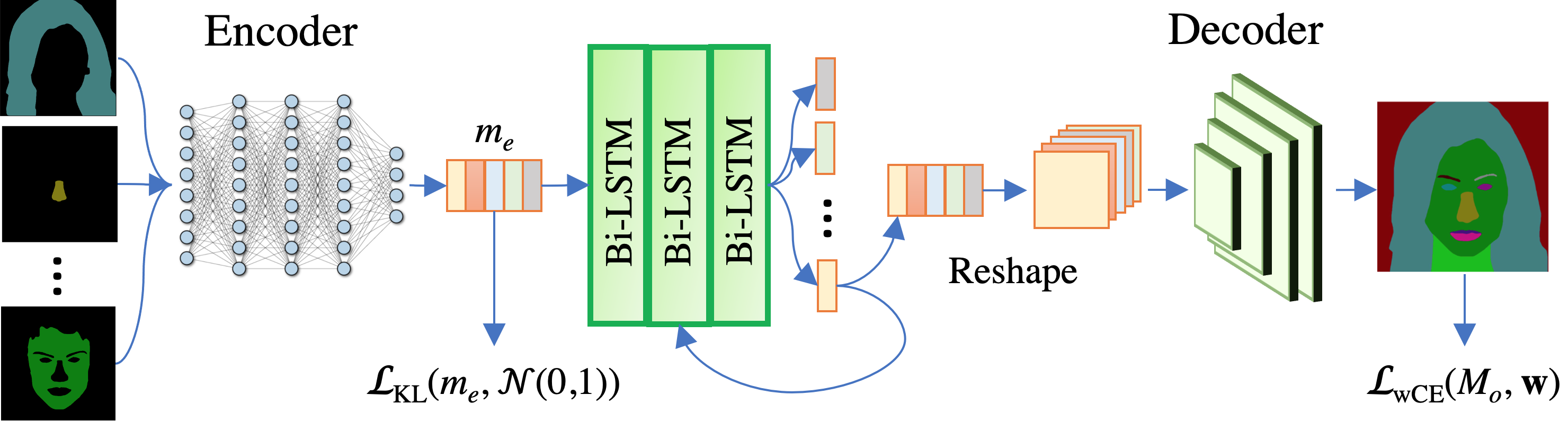}
    \caption{Proposed architecture: the segmentation mask $M \in [0,1]^{C \times 256 \times 256}$ is processed so that each channel (class) $c \in C$ is flattened and passes through a MLP encoder to obtain class-wise embeddings $m_e^c$ for each semantic class. The embeddings $m_e = [m_e^1, \cdots, m_e^C]$ then pass through a set of three bi-directional LSTM layers followed by a feed-forward block that learn relationships across the classes. The processed embeddings are finally reshaped to form a set of feature maps $m_d \in \mathbb{R}^{C \times 16 \times 16}$, and then fed to a convolutional decoder which outputs a new mask $M_o \in [0,1]^{C \times 256 \times 256}$. The model is trained with (1) a pixel-wise weighted cross-entropy loss ($\mathcal{L}_{wCE}$), and (2) a $KL$-divergence loss $\mathcal{L}_{KL}$ applied to the embeddings $m_e$ so to push them towards following a $\mathcal{N}(0,1)$ distribution, enabling their generation from noise or manipulation.}
    \label{fig:arch}
\end{figure}

\subsection{Architecture}
To account for the above challenges, we designed our proposed architecture (Fig.~\ref{fig:arch}) to have 4 main components: (1) an MLP $\mathcal{M}$ to independently encode the mask channels  into a latent representation $m_e$. This allows us to operate on the mask channels directly in the compressed space; (2) an LSTM-Feed Forward block $\mathcal{L}$ composed of three bi-directional LSTM layers to process the encoded mask channels $m_e^j$ and account for possible misalignments resulting from manipulating a semantic class, and a feed-forward block $\mathcal{F}$ to further re-arrange the processed mask encodings; (3) finally, a convolutional decoder $\mathcal{D}$ to reconstruct the complete semantic mask $M_o$.

\noindent\textbf{MLP Encoder.}
The encoder $\mathcal{M}$ is a simple MLP made up of three linear layers, each followed by a ReLU activation function. Each mask channel is first flattened so that the input mask has size $M \in \mathbb{R}^{C \times H^2}$ where $H = W = 256$ is the spatial size of the mask; each linear layer of the encoder has a hidden size of $256$, so that $m_e = [m_e^1, \cdots, m_e^C] = \mathcal{M}(M) \in \mathbb{R}^{C \times 256}$.

\smallskip
\noindent\textbf{Bi-directional LSTM Block.}
The bi-directional LSTM~\cite{schuster1997bidirectional} block was designed to process the encoded mask channels $m_e$ one after another, as if they were frames of a temporal sequence. The goal is that of correcting possible inconsistencies resulting from manipulating or generating a class embedding $m_e^c$, based on the information of the other classes. Intuitively, if we change the shape of a facial part in the mask \textit{e.g.} nose, the surrounding parts need to be adjusted so that the combined result looks realistic and artifact-free. One problem arising from using a recurrent module is that of choosing the order in which the channels are processed. Temporal sequences have a unique ordering implicitly defined by the time flow, whereas in our scenario there is no clear nor unique way of choosing the order by which processing the face parts, being them simply parts of a spatial layout. 
This motivated us to opt for the bi-directional variant of the LSTM; indeed, the latter processes the sequence in both directions (first to last, and last to first), so that each class embedding is influenced by all other classes, not only by the previously processed ones.
Each class embedding $m_e^c$ is thus processed, and provided as hidden state both for the subsequent $m_e^{c+1}$ and previous $m_e^{c-1}$ classes. In addition, differently from the standard use of LSTMs where only the last processed embedding keeps flowing through the network, we also store the embeddings at intermediate steps $m_e^c$. In doing so, once all the $C$ embeddings have been processed, we end up with the same number of $C$ embeddings, one for each class. 
Finally, following the same principle of~\cite{vaswani2017attention}, a feed-forward block composed of two linear layers equipped with GeLu~\cite{hendrycks2016gaussian} activation function is stacked after the LSTMs so to make the embeddings better fit the input to the decoder.

\smallskip
\noindent\textbf{Decoder.}
Finally, the convolutional decoder $\mathcal{D}$ is responsible for learning to reconstruct the segmentation mask from the $C$ embeddings resulting from the previous steps. In particular, the $C$ embeddings are reshaped into a set of $C$ feature maps $m_d \in \mathbb{R}^{C \times 16 \times 16}$. These are processed by 4 residual blocks, equipped with SiLU \cite{elfwing2018sigmoid} activation function and group normalization. The decoder outputs the reconstructed segmentation mask $M_o \in \mathbb{R}^{C \times 256 \times 256}$.

\subsection{Loss Functions}\label{subsec:loss}
The model is trained to self-reconstruct the input segmentation mask, without any other specific strategy to guide the manipulation process. The output mask is generated by minimizing a pixel-wise class prediction, using a cross-entropy loss. In particular, we used a weighted variant of the standard cross entropy $\mathcal{L}_{CE}$.
More in detail, we observed that the problem resembles a highly imbalanced classification problem; indeed, smaller parts such as, for face masks, the eyes or the nose, are significantly under-represented in the data \textit{i.e.} occupy a smaller number of pixels, with respect to larger parts such as skin or hair, ultimately weighing less in the overall loss computation. So, the weights are set considering this imbalance; smaller weights will be assigned to bigger parts, and bigger weights will be assigned to smaller parts. We calculate the weights $\textbf{w} = [w_0,\cdots,w_C]$ based on the overall training set statistics, in the following way:
\begin{equation}
    \textbf{w} = 1 - \frac{1}{NHW}\sum_N^i\sum_H^j\sum_W^kx_{c,i,j,k} \; \forall \; c \in C
\end{equation}
\noindent where $N$ is the number of samples in the training set, and $H$ and $W$ are the height and width of the semantic mask, respectively. Given that each of the mask channels can contain only one or zero values, this equation provides a series of $C$ weights that rank each of the semantic parts by their average size. The equation of the final weighted cross entropy $\mathcal{L}_{wCE}$ therefore becomes:

\begin{equation}
    \mathcal{L}_{wCE} = -\sum_x \textbf{w}(y(x))
 y(x)log(\hat{y}(x))
\end{equation}
\noindent where $y(x)$, $\hat{y}(x)$, and $\textbf{w}(y(x))$ are the ground-truth class labels, the predicted labels, and the weight for the ground-truth class at pixel $x$, respectively.

In addition to the weighted cross entropy, a KL-Loss $\mathcal{L}_{KL}$ is used to push the latent codes of each of the parts to have zero mean and unit variance and allow the generation process where a random latent code is sampled from $\mathcal{N}(0,1)$. Ultimately, the full loss utilized to train the model is:

\begin{equation}
    \mathcal{L} = \mathcal{L}_{wCE} + \lambda\mathcal{L}_{KL}
\end{equation}

\noindent where $\lambda$ is the KL weight and is set to 0.0005 in all the experiments.

\section{Experimental Results}
In this section, we report the results of an experimental validation.
We show both quantitative and qualitative results, in terms of reconstruction accuracy and different generation or manipulation tasks. In fact, despite our goal being that of performing editing of semantic masks at the class level, we also need to make sure the reconstruction process does not degrade the segmentation accuracy of the input masks and in turn compromise the subsequent image synthesis.

%\subsection{Dataset}
As dataset to train and test our model, we used the CelebAMask-HQ~\cite{lee2020maskgan}, which is composed by 30K high resolution face images (1024$\times$1024) along with the corresponding segmentation masks. Out of the 30K samples, 28K were used for training and 2K for testing.

\subsection{Reconstruction, Generation and Perturbation}
Given that no prior works addressed this particular problem, before analyzing the ability of the model to manipulate the mask parts, we compare our solution with some baseline architectural designs in terms of reconstruction accuracy, in a sort of an extended ablation study. 
Reconstruction results are reported in Table~\ref{tab:abl} in terms of pixel-wise classification accuracy (Acc) and Mean Intersection over Union (mIoU). In particular, the following configurations were explored: a simple encoder-decoder trained with the standard cross-entropy (row 1), the model with 1 or 3 standard LSTMs trained with standard cross entropy (rows 2 and 3), our final model with 3 standard LSTMs trained with the weighted cross entropy (row 4),  our final model with 3 bidirectional LSTMs trained with regular cross entropy (row 5), and the final architecture (bottom row). 

\begin{table}[!b]
    \centering
    \begin{tabular}{c|c|c}
        \textbf{Method} & mIoU $\uparrow$ & Acc $\uparrow$ \\
        \hline
        \hline
        w/o LSTM block & 68.49 & \textbf{94.24}\\
        \hline
        1 LSTM w/o Bidir. w/o weigthed CE & 68.15 & \underline{93.85} \\
        \hline
        3 LSTMs w/o Bidir. w/o weigthed CE & 67.35 & 90.91 \\
        \hline
        3 LSTMs w/o Bidir & 68.34 & 90.94 \\
        \hline
        3 LSTMs w/o weigthed CE & \underline{69.56} & 92.12 \\
        \hline
        \textbf{Ours} & \textbf{70.31} & 92.39 \\
        \hline
    \end{tabular}
    \caption{Reconstruction results comparing our solution with different baselines.}
    \label{tab:abl}
\end{table}

\begin{figure}[!t]
    \centering
    \includegraphics[width=\textwidth]{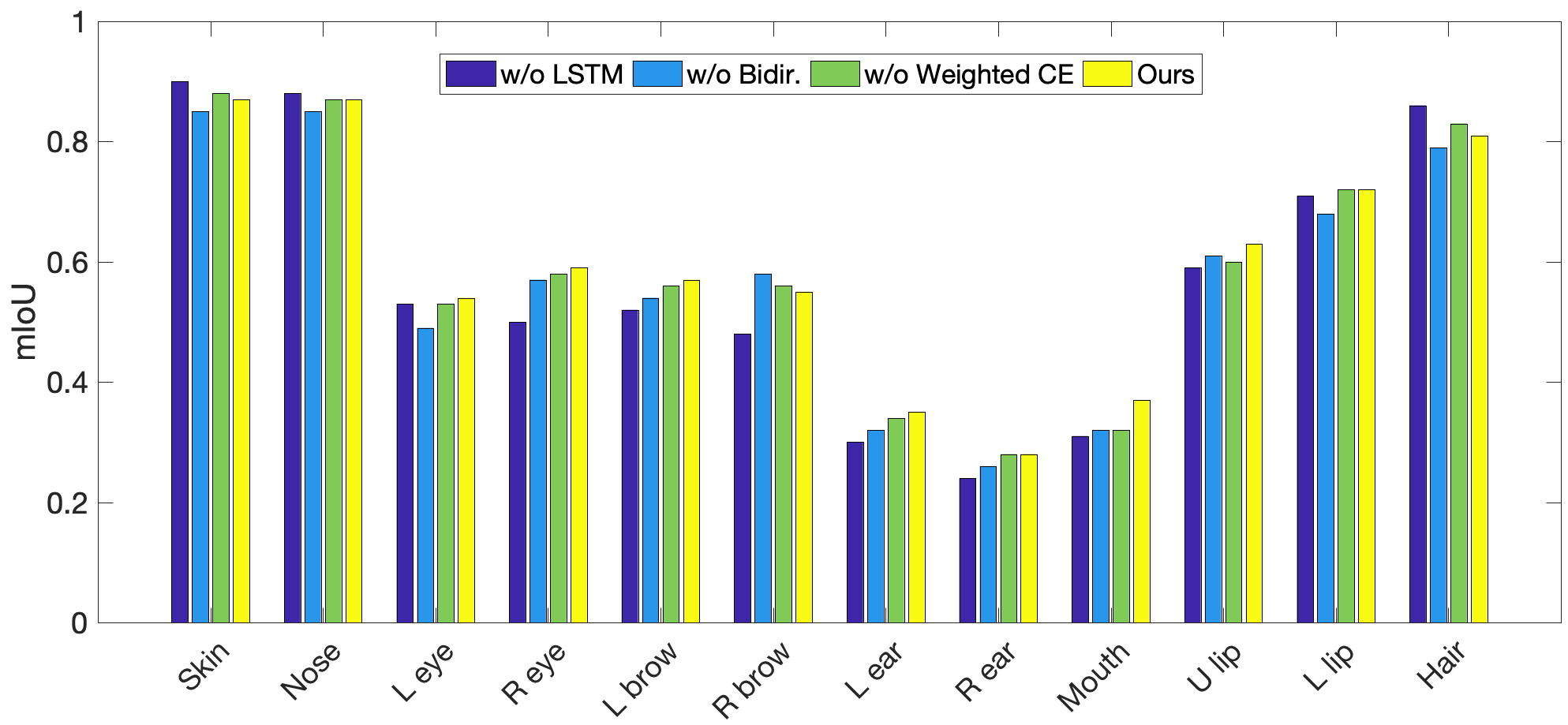}
    \caption{mIoU results per class.}
    \label{fig:plot}
\end{figure}

\begin{figure}
    \centering
    \includegraphics[width=0.7\textwidth]{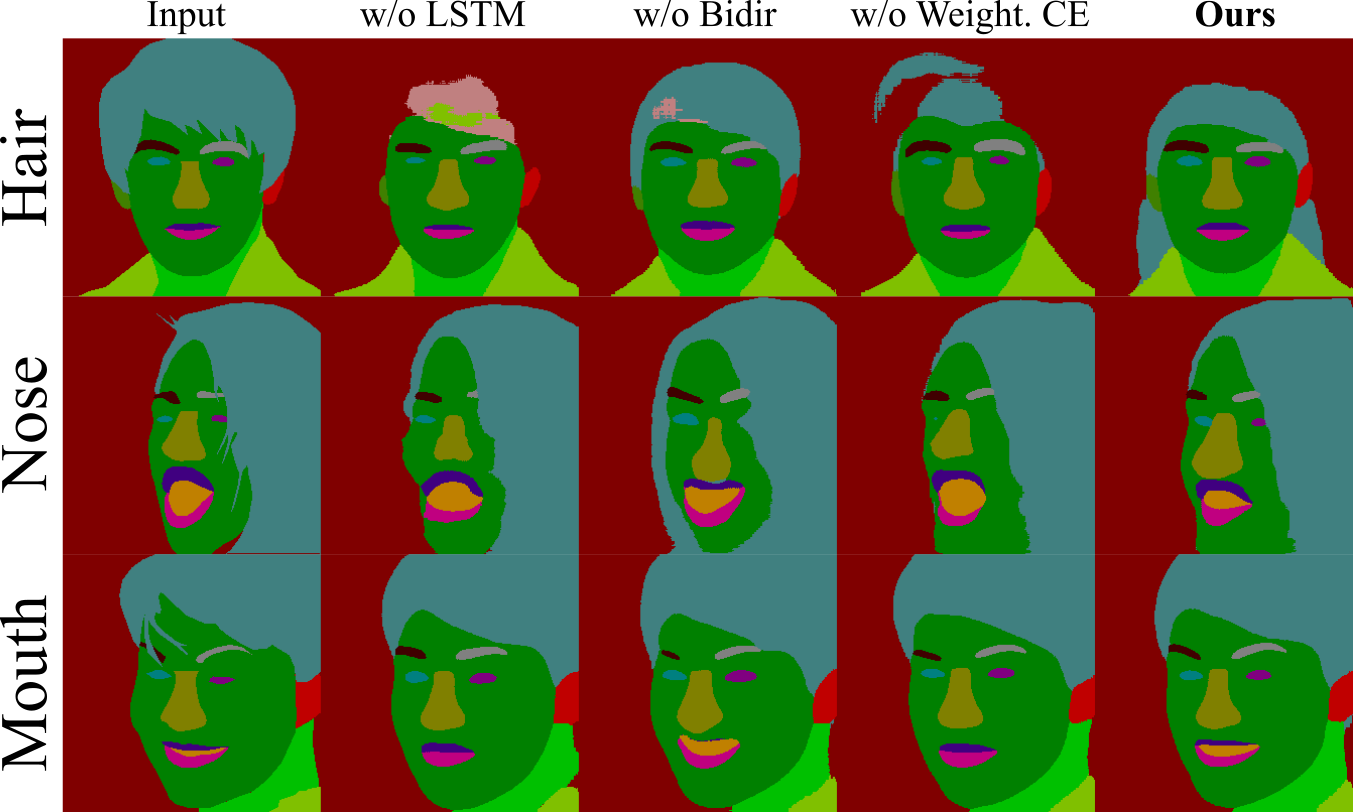}
    \caption{Qualitative results of different ablations experiments.}
    \label{fig:abl}
\end{figure}

Quantitatively, we observe a generally-high reconstruction accuracy in all the cases. The simplest architecture (w/o LSTM and weighted CE) achieves the highest accuracy but lower mIoU. A visual inspection of the results suggests that the the additional processing due to the LSTM block induces a slight smoothing of high-frequency details such as the hair contour. This is caused by the compression of each semantic part in the encoding phase, and also by the Bi-directional LSTM block pass which makes more difficult for the decoder to exactly reproduce the corresponding input. This hypothesis is supported if looking at the results obtained with either 1 or 3 LSTM layers; indeed the two measures decrease when stacking more LSTM layers. On the other hand though, we will show (Fig.~\ref{fig:abl}) that removing such layers severely compromises the manipulation ability. Nevertheless, when comparing configurations including 3 LSTM layers, our final architecture scores the highest accuracy. In particular, it obtains the highest mIoU, which indicates the overall shape and spatial arrangement of parts is best preserved. This is also supported by the results in Fig. \ref{fig:plot}, which shows per class mIoU results of different configurations. Indeed, even though the configuration w/o LSTM tends to perform better with bigger parts like skin or hair, our final architecture manages to push the quality of the smaller parts up thanks to the combination of bidirectional LSTMs and weighted cross entropy loss, resulting in an overall better mIoU.

\begin{figure}[!h]
    \centering
    \includegraphics[width=\textwidth]{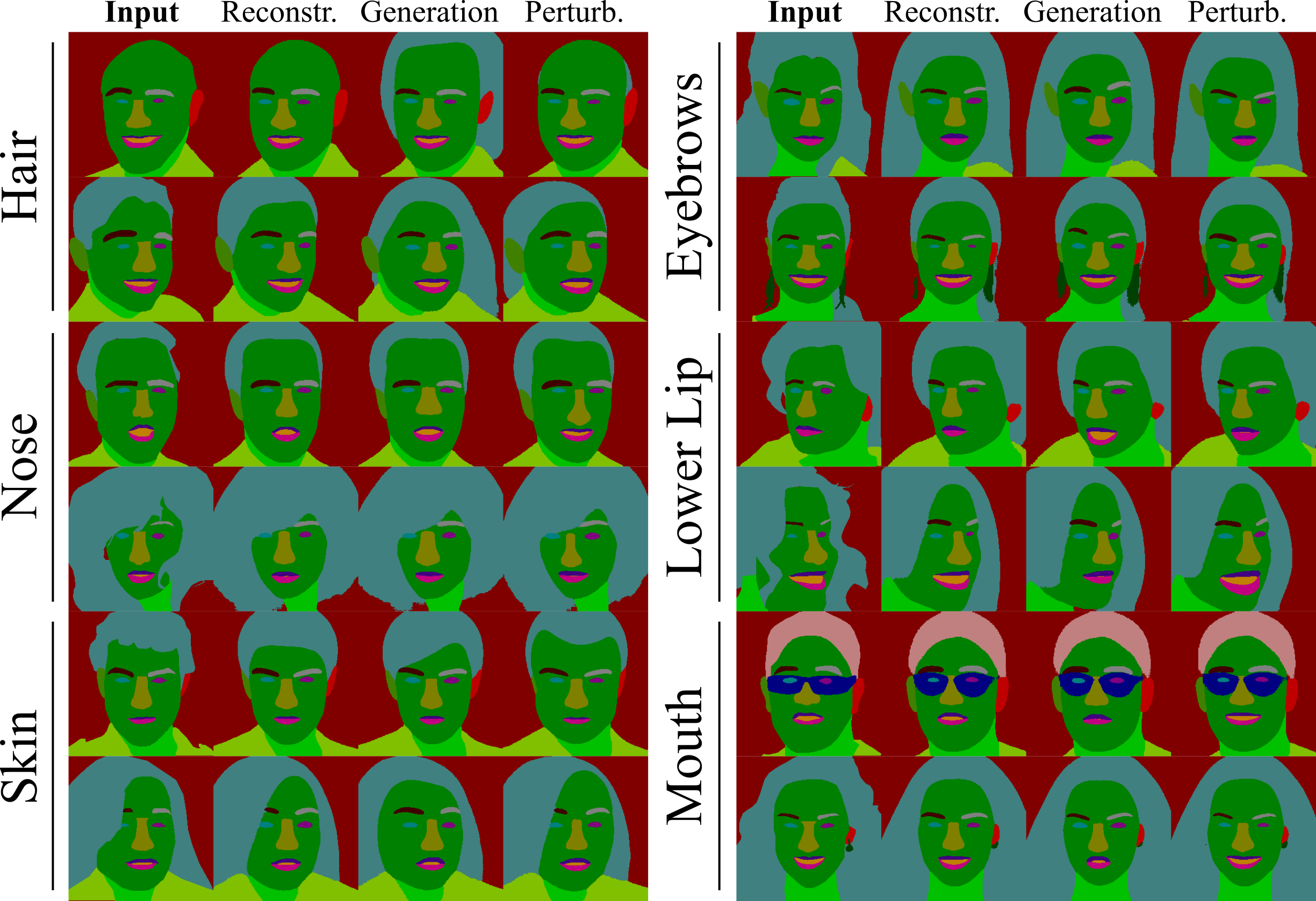}
    \caption{Results for reconstruction, generation and perturbation of different mask parts.}
    \label{fig:res}
\end{figure}

In Fig.~\ref{fig:res} some results for both reconstruction, generation and perturbation of different parts in the semantic masks are presented. More in detail, we refer to \textit{generation} when a novel latent code drawn from the normal distribution $\hat{m}_e^j \sim \mathcal{N}(0,1)$ is substituted to its encoded counterpart $m_e^j$ and passed to the bi-directional LSTM block in order to generate a particular part $c$. On the other side, we refer to \textit{perturbation} when a random noise vector drawn from the normal distribution $z \sim \mathcal{N}(0,1)$ is added to an existing latent code \textit{i.e.} $\hat{m}_e^j = m_e^j + z$. Indeed, in the latter, usually the shape of the generated parts is more similar to the original input, while in the first case the generated shape can be (and usually is) completely different.

Regarding reconstruction, we can see how the proposed method manages to maintain the overall shape of the semantic mask parts, supporting the results in Table \ref{tab:abl}. Nevertheless, as discussed above, a certain degree of smoothing in the results can be noted. This represents a minor limitation of the current proposal.
On the other side, results when generating parts from scratch, or by perturbing an existing latent code, are impressive. Our method is not only able to generate realistic parts independently from one another, but also, thanks to the recurrent part of the model, is able to adapt the shape of the parts surrounding the one that is being generated in order to produce  a realistic final result. This can be particularly appreciated for example when perturbating the nose latent code in Fig \ref{fig:res} in the third row: indeed, the nose is made longer by the perturbation and as a consequence the mouth is deformed accordingly.

Finally, in Fig. \ref{fig:abl} we show some qualitative results to prove that the final architecture is indeed better in the generative task which is the main purpose of this paper. Starting from the top, it is clear how when generating hair the proposed model is much more capable of producing a realistic results without generating undesired classes (like the pink part in the model without LSTM). Then, in the second row, is proved how our model is much better at rearranging all the semantic parts in order to create a realistic mask with a newly generated part. Finally, in the last row, we can see how the mouth part is generated correctly by almost every configuration, but, at the same time, our model is able to generate much more varied and diverse results.

\subsection{Interpolation}

\begin{figure}[h]
    \centering
    \includegraphics[width=0.9\textwidth]{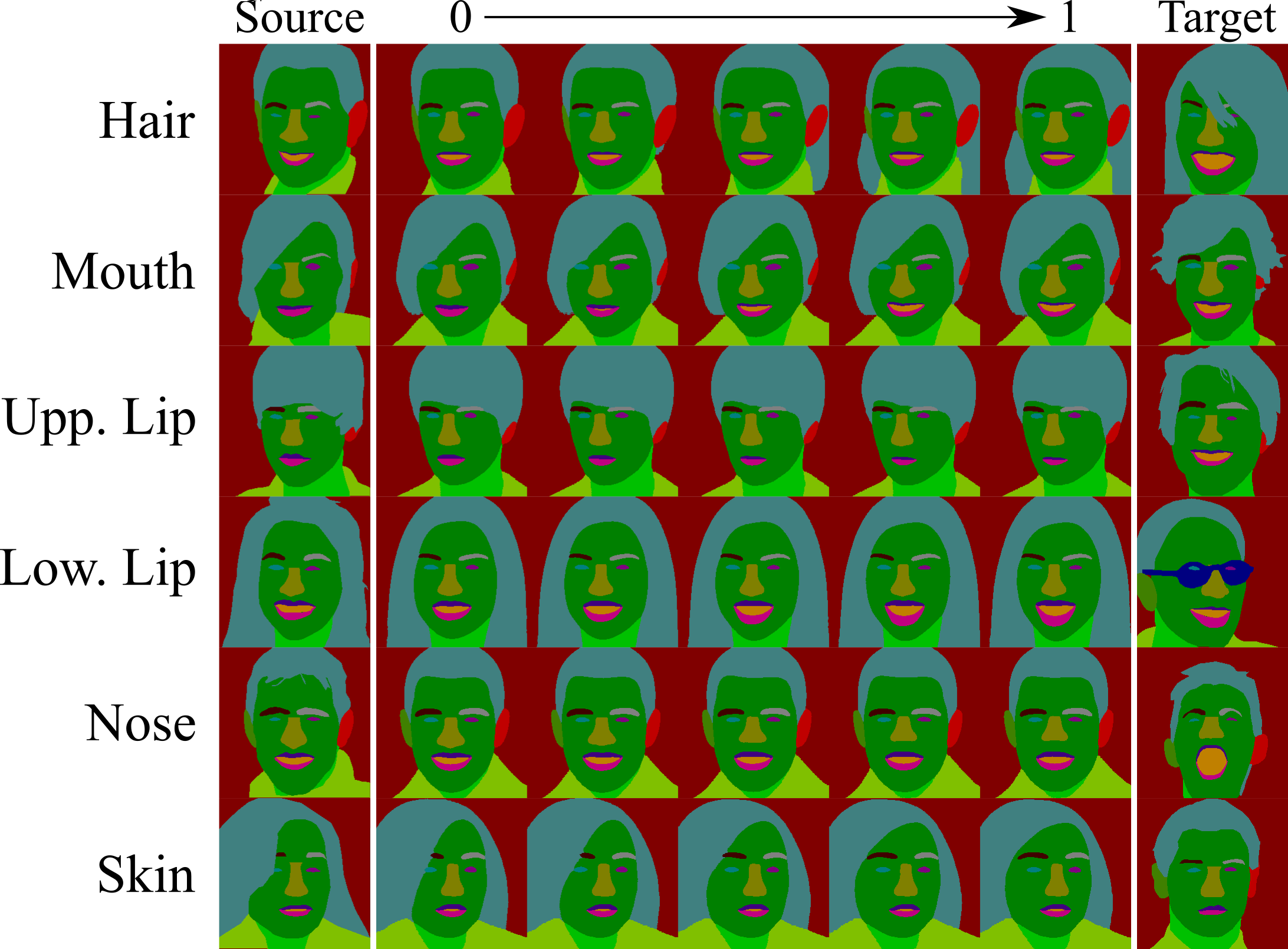}
    \caption{Interpolation results of different parts taken from a source and target mask (first and last columns, respectively). Values of the interpolation factor $\alpha$ go from 0 to 1, where 0 means no interpolation.}
    \label{fig:int}
\end{figure}

In Fig. \ref{fig:int} interpolation results are presented. Interpolation is done by choosing a part $c$ from a source and a target mask and merging together the corresponding latent vectors using an interpolation factor $\alpha$. More in detail, the interpolation equation is the following:

\begin{equation}
    m^{int}_c = \alpha \cdot m^{t}_c + (1-\alpha)\cdot m^{s}_c
\end{equation}

\noindent where $m^{t}_c$ and $m^{s}_c$ are the latent codes of the part $c$ of the target and source images, respectively. In addition, $\alpha$ = 0 is equal to reconstructing the source image, while $\alpha$ = 1 represents a sort of ``face part swapping'', that is a specific face part is swapped from a target face to a source one.

Indeed, it is evident how the KL loss, that pushes the latent codes to have almost zero mean and unit variance, allows to easily interpolate every mask part. In particular, while increasing the interpolation factor $\alpha$, the shape changes continuously. The only previous method that we are aware of capable of performing a similar task is MaskGAN~\cite{lee2020maskgan}; however, MaskGAN can only perform global mask interpolations, and can not independently manipulate individual parts.

%Table \ref{tab:abl} presents a series of quantitative ablations studies of different versions of the proposed architecture. More in detail, the numbers refers to the quality in the reconstruction of the generated samples when no single-part generation is happening. The reason for this experiments is to prove that the architecture design, that is aimed at allowing automatic generation of semantic parts, does not hurt the reconstruction capability of the model.

%Numbers in the tables refer to the mean intersection over union (mIoU) and pixel-by-pixel accuracy (Acc) in the generated samples compared with the groud truth images.

%Looking at the table, it is evident how the best accuracy is held by the model without LSTMs while the proposed model reaches the best mIoU. This can be explained by the fact that, in the first case, no additional modules between encoder and decoder allow to reach a better precision in reproducing shapes without smoothing, while, on the other hand, the final architectures is more capable in positioning all the semantic parts correctly in the image, resulting in a better mIoU.

\subsection{Semantic Image Synthesis with Shape Control}

\begin{figure}[!t]
    \centering
    \includegraphics[width=0.9\textwidth]{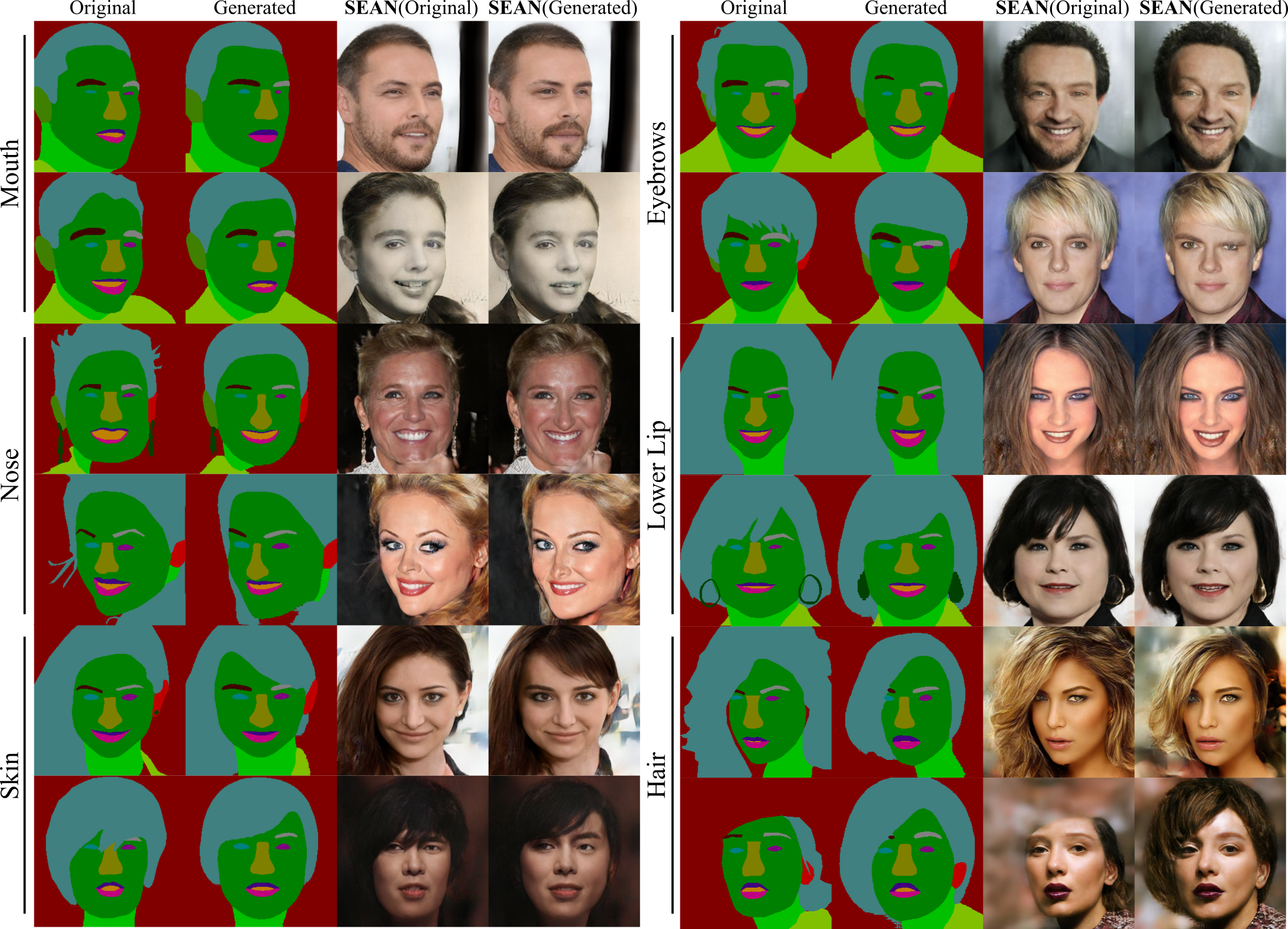}
    \caption{RGB results obtained when feeding the original mask and a mask automatically edited with our method to a SIS model.}
    \label{fig:rgb}
\end{figure}

In this section we qualitatively show results for the main purpose of our model, that is equipping SIS generators with a module to enable automatic shape control. In Fig.~\ref{fig:rgb}, several mask with automatically generated parts are fed to a state-of-the-art SIS model in order to produce new and diverse face images. 
We chose to use the SEAN~\cite{park2019semantic} generator to this aim because SEAN can very precisely control the image generation thanks to its semantic region-adaptive normalization layers.
Previous to our proposal, the editing of masks could only be done manually. Results in Fig.~\ref{fig:rgb} clearly show that, provided a generator that is accurate enough to handle local shape changes, the shape of the generated faces can be automatically edited by means of our solution. This paves the way to a very efficient way of employing SIS models, for example, for data augmentation which can be very helpful for task like re-identification, classification or detection. 

\section{Conclusion}
In this paper, we introduced the problem of automatic manipulation of semantic segmentation masks, and presented a preliminary novel architecture to achieve this goal, with a specific application to face part editing. %It consists in an encoder-decoder structure with a Bi-directional LSTM Block in between. The model is trained using a combination of reconstruction loss and KL divergence.
The proposed system is able do generate or manipulate any semantic part by just feeding random noise to the LSTM block in the place of the latent representation of the corresponding part. We show the efficacy of our architecture through a series of quantitative and qualitative evaluations. Even if we observed the tendency of smoothing the shapes of the generated results, still our method is able to generate realistic semantic parts, and can be readily used in combination with potentially any SIS models so to generate a virtually infinite number of RGB results.

Finally, we believe there is still large room for improvements. For example, extending the proposal to different scenarios with less constrained objects layout or more classes would represent a valuable feature for a SIS model. Also, currently, the shape manipulation is not controlled, meaning that it is not yet possible to generate parts with a specific shape or attributes, \textit{e.g.} long nose or curly hair. All the above are features that we plan to investigate in future works.

\section{Acknowledgments}
This work was supported by PRIN 2020 “LEGO.AI: LEarning the Geometry of knOwledge in AI systems”, grant no. 2020TA3K9N funded by the Italian MIUR.
%
% ---- Bibliography ----
%
% BibTeX users should specify bibliography style 'splncs04'.
% References will then be sorted and formatted in the correct style.
%
\bibliographystyle{splncs04}
\bibliography{bibliography}
\end{document}